\RequirePackage{fix-cm}
\RequirePackage{amsmath} 
\documentclass[twocolumn]{svjour3}           
\smartqed  
\usepackage{graphicx}  
\usepackage{picinpar}  

\journalname{Current Robotics Reports} 

\usepackage{graphicx}            
\usepackage{amssymb}             
\usepackage{amsfonts}            
\usepackage{enumitem}            
\usepackage{multirow}            
\usepackage{siunitx}             
\usepackage{url}                 
\usepackage{xspace}              
\usepackage[T1]{fontenc}         
\usepackage[hidelinks]{hyperref} 

\usepackage{pdflscape}              
\usepackage{afterpage}
\usepackage{booktabs}
\usepackage{caption}
\usepackage[misc]{ifsym}         

\graphicspath{{images/}}
\DeclareGraphicsExtensions{.pdf,.png,.jpg,.jpeg}

\newcommand{\figlabel}[1]{\label{fig:#1}}
\newcommand{\tablabel}[1]{\label{tab:#1}}

\newcommand{\figref}[1]{Fig.~\ref{fig:#1}\xspace}
\newcommand{\tabref}[1]{Table~\ref{tab:#1}\xspace}

\begin{document}

\title{Bipedal Humanoid Hardware Design: A Technology Review}

\author{Grzegorz Ficht \and Sven Behnke}

\institute{Grzegorz Ficht, ficht@ais.uni-bonn.de \Letter\\
Sven Behnke, behnke@ais.uni-bonn.de\\
All authors are at:\at
Rheinische Friedrich-Wilhelms-Universit{\"a}t Bonn\\
Friedrich-Hirzebruch-Allee 8, 53115 Bonn\\
}

\date{Received: date / Accepted: date}

\maketitle

\begin{abstract}~\newline
\textbf{Purpose of Review} As new technological advancements are made, humanoid robots that utilise them are being designed and manufactured. For optimal design choices, 
a broad overview with insight on the advantages and disadvantages of available technologies is necessary. This article intends to provide 
an analysis on the established approaches and contrast them with emerging ones. \newline
\textbf{Recent Findings} A clear shift in the recent design features of humanoid robots is developing, which is supported by literature. 
As humanoid robots are meant to leave laboratories and traverse the world, compliance and more efficient locomotion is necessary.
The limitations of highly rigid actuation are being tackled by different research groups in unique ways. Some focus on modifying the kinematic structure,
while others change the actuation scheme. With new manufacturing capabilities, previously impossible designs are becoming feasible.\newline
\textbf{Summary} A comprehensive review on the technologies crucial for bipedal humanoid robots was performed. Different mechanical concepts have been discussed,
along with the advancements in actuation, sensing and manufacturing. The paper is supplemented with a list of the recently developed platforms along with a selection of their specifications.

\keywords{Humanoid Robotics \and Kinematic Structure \and Actuation \and Sensors \and 3D-printing \and Manufacture}
\end{abstract}

\section{Introduction}

Humanoid Robotics has come a long way in its development, with progress accelerating in its every branch by the years. 
What started as basic, yet fascinating automata has quickly grown into a global interdisciplinary research field in 
just half of a century. The intertwined dependence of theoretical concepts and available technologies is constantly 
shaping the evolution of Humanoid Robotics. Current humanoid robots are not the result of a single scientific breakthrough, 
but rather an accumulation of small incremental achievements in the respective utilised technologies. 
Humanoids are not the sole beneficiary though, as the relation is mutual. More advanced platforms resulted in increased 
application complexity, which in turn facilitates the creation of novel technological solutions. Due to these advancements, 
humanoid robots come in different shapes, sizes and functionality (see \figref{robots_img}). Much work has been done since the early 2000s, where humanoids
were more in the fictional domain, with only few capable robots available~\cite{behnke2008humanoid}. With limited actuation 
capabilities, only smaller platforms were accesible for many researchers. Size is a limiting factor, as smaller robots are unable to meaningfully interact 
in an environment meant for humans, which is the ultimate goal of developing these robots. On the other hand, larger robots require 
disproportionally more actuation power---making the task of building such much more difficult~\cite{froese2006cube}. As in 
nature~\cite{usherwood2020fastest}, mediating size and performance is crucial for achieving an optimal middle-ground.
For all intents and purposes, this review focuses on robots that aim to mimic the bipedal mobility of a human, and the 
recent advancements in technology that enable it.

\newpage

\begin{landscape}
\begin{figure}
\parbox{\linewidth}{
\includegraphics[width=1.0\linewidth]{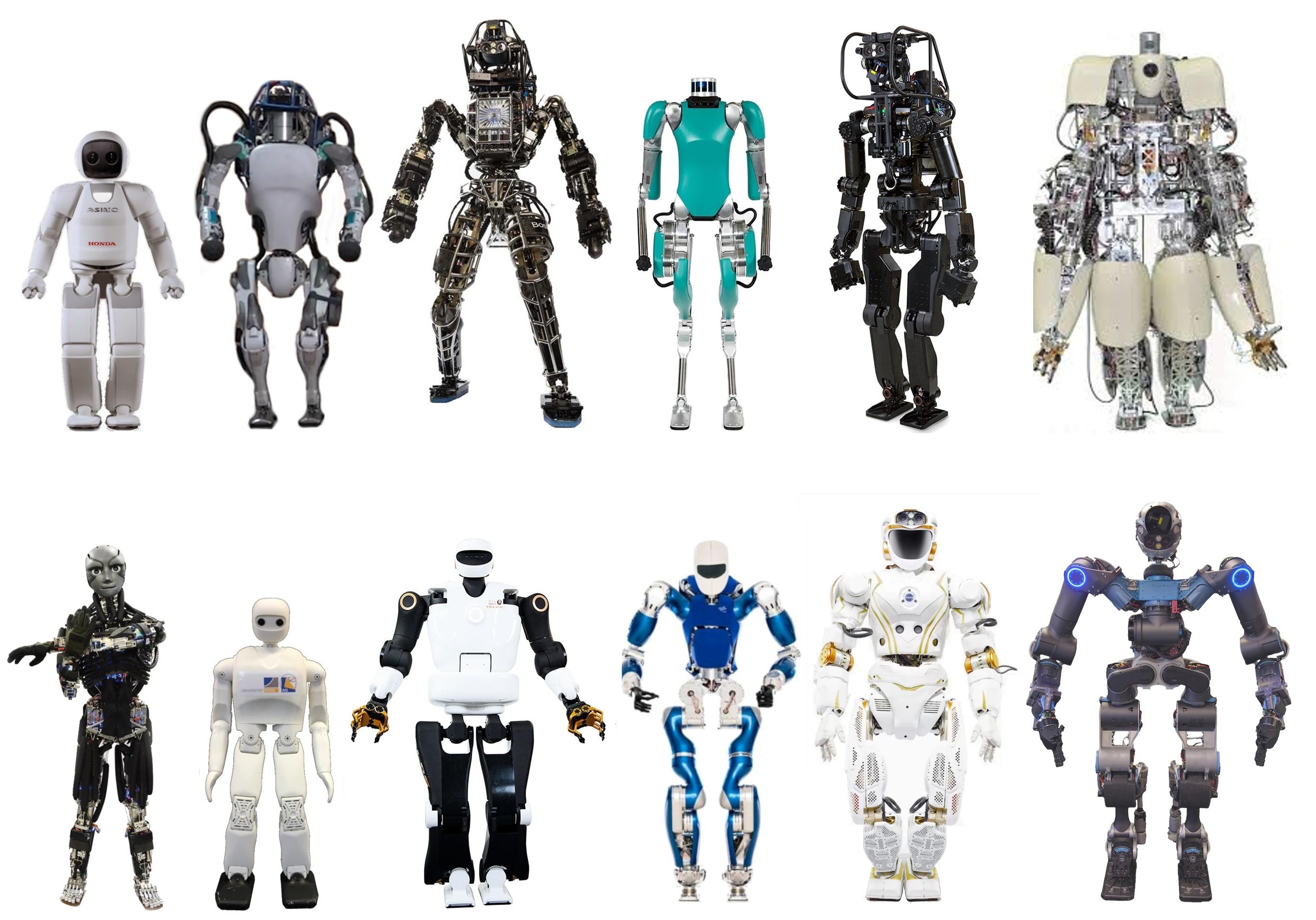}} 
\captionsetup{justification=centering,margin=2cm}
\caption{\footnotesize Most recently developed bipedal humanoid robots with a full humanoid body plan, from left to right, top to bottom: 
\newline Asimo, Atlas, Atlas-Unplugged, Digit, HRP-5P, Hydra, Kengoro, NimbRo-OP2X, TALOS, Toro, Valkyrie, WALK-MAN.}
\figlabel{robots_img}
\vspace{-3ex} 
\end{figure}

\end{landscape}

\newpage
\begin{landscape}


\begin{table}
\renewcommand{\arraystretch}{1.15}
\caption{\footnotesize\vspace{0.7ex}Most recently developed bipedal humanoid robots with a full humanoid body plan.\vspace{-0.7ex}}
\tablabel{robots}
\centering
\begin{tabular}{c c c c c c c c c c}
\hline
\multirow{2}{*}{Name} & Height & Weight & \multirow{2}{*}{Actuation} & No. of & \multirow{2}{*}{Sensing} & \multirow{2}{*}{Manufacture} & \multirow{2}{*}{Year} & Tentative \\
& (cm) & (kg) & & actuators & & & & price\\
\hline
\\
\multirow{2}{*}{Asimo (2011 model)} & \multirow{2}{*}{130} & \multirow{2}{*}{48} & Electric & \multirow{2}{*}{57} & Joints: position & Magnesium & \multirow{2}{*}{2011} & \multirow{2}{*}{\SI{2500000}{USD}} \\
 &  &  & Harmonic Drive &  & IMU, 2x F/T, Camera & alloy &  &  \\
\\
\multirow{2}{*}{Atlas (Next Generation)} & \multirow{2}{*}{150} & \multirow{2}{*}{75} & Hydraulic & \multirow{2}{*}{28} & Joints: position, force & Metal,  & \multirow{2}{*}{2016} & \multirow{2}{*}{N/A} \\
 &  &  & Servo-valves &  & Lidar, Stereo vision & 3D-printed &  &  \\
\\
\multirow{2}{*}{Atlas-Unplugged} & \multirow{2}{*}{188} & \multirow{2}{*}{182} & Hydraulic & \multirow{2}{*}{30} & Joints: position, force & Aluminium & \multirow{2}{*}{2015} & \multirow{2}{*}{\SI{2000000}{USD}} \\
 &  &  & Servo-valves &  & Lidar, Stereo vision & Titanium &  &  \\
\\
\multirow{2}{*}{Digit} & \multirow{2}{*}{155} & \multirow{2}{*}{42.2} & Electric & \multirow{2}{*}{16} & Joints: position & Aluminium, milled & \multirow{2}{*}{2019} & \multirow{2}{*}{\SI{250000}{USD}} \\
 &  &  & Cycloid Drive &  & IMU, Lidar, 4x Depth Cam. & Carbon fiber &  &  \\
\\
\multirow{3}{*}{HRP-5P} & \multirow{3}{*}{183} & \multirow{3}{*}{101} & \multirow{2}{*}{Electric} & \multirow{3}{*}{37} & Joints: position & \multirow{2}{*}{Metal} & \multirow{3}{*}{2018} & \multirow{3}{*}{N/A} \\
 &  &  & \multirow{2}{*}{Harmonic Drive} &  & 4x F/T, IMU, Lidar & \multirow{2}{*}{(unspecified)} &  &  \\
 &  &  &  &  & Stereo Vision &  &  &  \\
\\
\multirow{2}{*}{Hydra} & \multirow{2}{*}{185} & \multirow{2}{*}{135} & Hydraulic & \multirow{2}{*}{41} & Joints: position, force & Aluminium & \multirow{2}{*}{2016} & \multirow{2}{*}{N/A} \\
 &  &  & EHA &  & IMU, 2x F/T, Lidar, Stereo & milled &  &  \\
\\
\multirow{2}{*}{Kengoro} & \multirow{2}{*}{167} & \multirow{2}{*}{55.9} & Electric & \multirow{2}{*}{106} & Joints: position, tension & Aluminium & \multirow{2}{*}{2016} & \multirow{2}{*}{N/A} \\
 &  &  & Muscle /w Tendons &  & IMU, 2x F/T, Stereo Vision & 3D-printed &  &  \\
\\
\multirow{2}{*}{NimbRo-OP2(X)} & \multirow{2}{*}{135} & \multirow{2}{*}{19} & Electric & \multirow{2}{*}{34} & Joints: position & PA12 Nylon & \multirow{2}{*}{2017} & \multirow{2}{*}{\SI{25000}{EUR}} \\
 &  &  & DC Servo-motors &  & IMU, Stereo Vision & 3D-printed &  &  \\
\\
\multirow{2}{*}{TALOS} & \multirow{2}{*}{175} & \multirow{2}{*}{95} & Electric & \multirow{2}{*}{32} & Joints: position, torque & Metal & \multirow{2}{*}{2017} & \multirow{2}{*}{\SI{900000}{EUR}} \\
 &  &  & Harmonic Drive &  & IMU, RGBD camera & (unspecified) &  &  \\
\\
\multirow{2}{*}{Toro} & \multirow{2}{*}{174} & \multirow{2}{*}{76.4} & Electric & \multirow{2}{*}{39} & Joints: position, torque & Aluminium & \multirow{2}{*}{2014} & \multirow{2}{*}{N/A} \\
 &  &  &  Harmonic Drive &  & 2x IMU, RGB\&D cameras & milled &  &  \\
\\
\multirow{2}{*}{Valkyrie} & \multirow{2}{*}{187} & \multirow{2}{*}{129} & Electric & \multirow{2}{*}{44} & Joints: position, force, torque & Metal & \multirow{2}{*}{2013} & \multirow{2}{*}{\SI{2000000}{USD}} \\
 &  &  & SEA &  & 7x IMU, 2xF/T, Multiple cameras & (unspecified) &  &  \\
\\
\multirow{3}{*}{WALK-MAN} & \multirow{3}{*}{191} & \multirow{3}{*}{132} & \multirow{2}{*}{Electric} & \multirow{3}{*}{29} & Joints: position, torque & \multirow{2}{*}{Aluminium} & \multirow{3}{*}{2015} & \multirow{3}{*}{N/A} \\
 &  &  & \multirow{2}{*}{SEA} &  & 2x IMU, 4x F/T, Lidar & \multirow{2}{*}{milled} &  &  \\
 &  &  &  &  & Stereo Vision &  &  &  \\

\hline

\end{tabular}
\end{table}

\end{landscape}

\section{Mechanical Structure}

The defining feature of any bipedal humanoid is its kinematic structure. Initially, the limited technology allowed 
only for a simplified representation of the humanoid form, often having just the legs. This is not a disadvantage 
as the design concept might prefer simpler forms, which allow for certain assumptions to be made and exploited in 
the control part~\cite{shin2019mechanistic}. Perfect examples of this approach are the MIT 3D Biped~\cite{playter1992control} and ATRIAS~\cite{hubicki2016atrias}. The extending leg with a 
rotating hip is enough to perform the 3D positioning function without having a human-like form. Both robots 
uniquely embody the Spring-Loaded Inverted Pendulum (SLIP) template for control. The MIT 3D Biped does so with 
linear thrusters, while ATRIAS has a pair of rotary actuators connected to the anterior and posterior thigh members 
of a 4-bar parallel linkage through spring plates. Both structures achieve a hip-concentrated mass and low-inertia legs, 
however due to the antagonistic actuation regime stemming from the leg design of ATRIAS, a significant 
amount of energy is wasted~\cite{abate2018mechanical}. Although impressive, these robots are chiefly used as demonstrators, 
for exploration of dynamic bipedal locomotion. 

For more general applications, robots with a full humanoid body plan have been built. Prominent examples of humanoids 
such as ASIMO~\cite{shigemi2018asimo}, the HRP series~\cite{yokoi2004experimental}~\cite{hrp2}~\cite{kaneko2008humanoid}~\cite{kaneko2011humanoid}~\cite{Kaneko2009}~\cite{kaneko2019humanoid}, 
HUBO~\cite{park2007mechanical}, REEM-C~\cite{robotics2015reem}, and TALOS~\cite{stasse2017talos} are all similar in terms of kinematics. 
The legs are composed of a three-degree-of-freedom (DoF) hip to simulate a spherical joint, one DoF for knee bending and two DoF for an 
ankle ball joint. As such, six actuators are enough to provide roughly the same form and functionality as a human leg. 
This structure has the added benefit of enabling a closed-form solution for the inverse kinematics. More joints (e.g. a toe) can 
be introduced to avoid singularities, joint limits, or achieve a specific configuration~\cite{tolani2000real}.
The mentioned robots utilise rotary joints with servo-motors in a serial configuration~(\figref{joints}.a.). The structure is well-established 
in the scientific community and has been thoroughly investigated, as evidenced by the number of robots that incorporate it. 
Its strength lies in the simplified mechanics and control, 
which comes at the drawback of limited capability. As the rotary actuators are typically placed directly at the joint, the ones closer to the origin of the chain 
need to carry the ones lower in the chain. This has two-fold consequences in backlash or elasticity accumulation from each joint and increased leg inertia.
Both contribute to positioning inaccuracies that the control scheme must compensate for.

Decreased limb rigidity and increased leg inertia can be substantial, which impacts the overall 
dynamic performance of the system and sets higher requirements for actuators in terms of quality and power output.
A remedy to this issue is to attach the actuators off-axis, closest to the root of the link and 
using a lightweight coupling or transmission (e.g. synchronous belts) as in an earlier version of ASIMO~\cite{hirose2007honda}~(see \figref{joints}.b.). Placing the actuators off-axis leads to 
interesting joint designs where parallel linkages can be involved. Parallel mechanisms exhibit higher stiffness
due to the mechanical drive coupling, which averages out the actuation error. The downside 
of parallel solutions is a limited workspace due to the matings of the links as well as an increased mechanical 
and computational complexity. Some mechanisms can be incorporated for specific joints, and some for whole legs.
A Stewart platform type leg~\cite{ceccarelli2020parallel} might provide benefits in terms of 
loading capabilities, but the limited range of motion and lack of a knee joint do not translate to human-like capabilities~\cite{tazaki2019parallel}. 

\begin{figure}[!b]
\vspace{-2ex}
\centering{\includegraphics[width=0.99\linewidth]{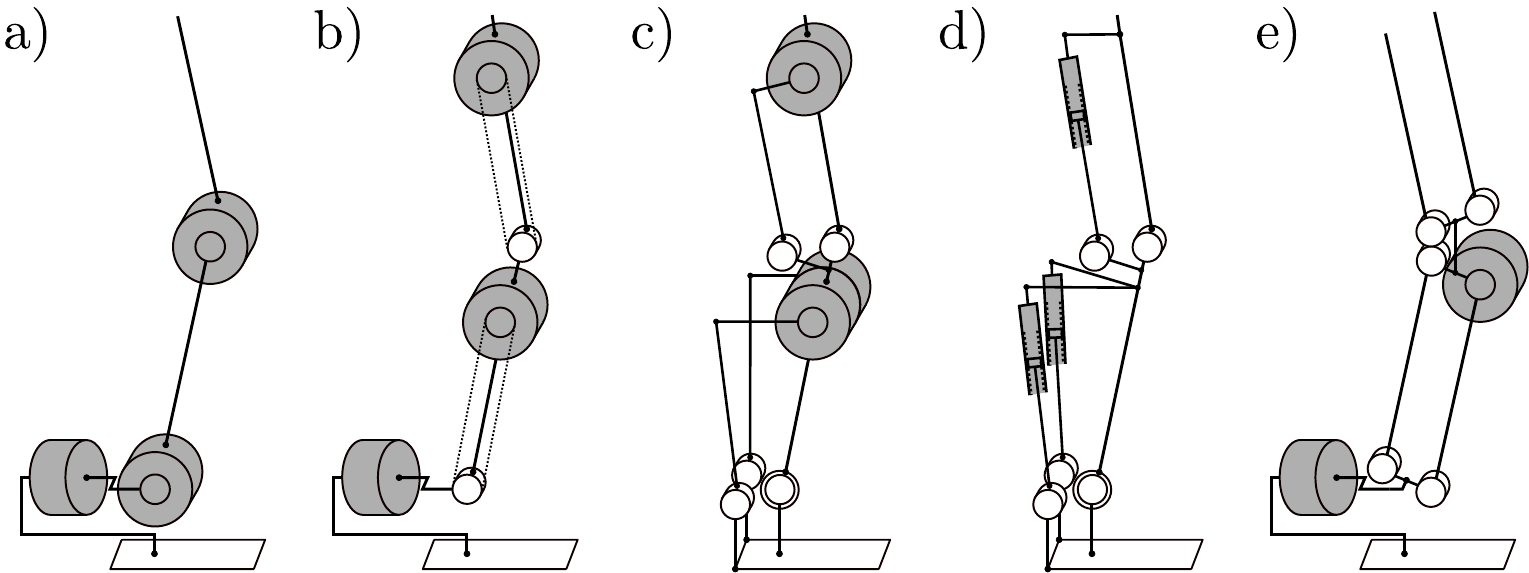}}
\caption{\footnotesize Configuration examples of lower leg design: a) on-axis serial mechanism, b) off-axis serial mechanism, 
c) crank-lever parallel mechanism, d) linear actuator parallel mechanism, e) mixed serial/parallel mechanism. Grey components represent actuators.}
\figlabel{joints} 
\end{figure}

In terms of human-like joints, A 3-DoF hip is least to benefit from the incorporation of a parallel structure 
as it is the root of the leg. Decreasing the inertia just marginally requires a significant investment in mechanical 
development~\cite{lee2014design}. Due to these mechanical intricacies, the yaw-ing in the hip is still usually performed 
with a separate rotary actuator~\cite{mghames2019spherical}, with its backlash marginally impacting the positioning of the whole leg.
The hip ab/adduction as well as flexion and extension can be performed with two actuators,
where the sum of their movements amounts to pitching, while differential movement results in rolling the hip~\cite{lahr2013early}.
As a high range of motion in the hip is desirable, only few robots up to date have incorporated parallel linkages in the hip,
such as the recent RHP-2~\cite{kakiuchi2017development}.
Implementation of a parallel knee joint is comparably easy, as there is usually sufficient 
space in the thigh link to place the actuator and transmission. The possible options involve employing 
either a rotary actuator in a crank-lever mechanism as seen in WALK-MAN~\cite{negrello2016walk} and the newer
ASIMO~\cite{kamioka2019push} or a linear actuator~\cite{lohmeier2006leg}, which both pull and push the shank link. In combination with a non-trivial linkage mechanism, 
the required knee torque can be significantly reduced, leading to a higher torque-to-weight ratio and thus, higher performance~\cite{tomishiro2019design}.
A 2-DoF ankle joint provides the most benefits of using parallel linkages, as the increased stiffness complements
a statically stable upright posture without much need for torque expenditure. Similar to the thigh, the shank provides ample space
for housing the actuators and couplings, which allow for pitching and rolling the ankle around a Cardan joint. 
As a result, several compact and elegant solutions have been produced. This includes the crank-lever mechanism found in ASIMO~\cite{asimopatent}, TORO~\cite{englsberger2014overview}, and CogIMon~\cite{zhou2018comprehensive}~(\figref{joints}.c.),
as well as the linear actuator scheme~\cite{lohmeier2006leg} \cite{hancock2020adaptive} \cite{otani2013new}~(\figref{joints}.d.).  

Apart from joint-specific solutions, few researchers took a different approach to the kinematic structure.
Gim et al.~\cite{gim2018design} designed a hybrid serial-parallel leg, where five
of the six leg actuators are located in the hip. This greatly reduces the leg inertia and in turn the required 
joint velocities and torques, which are also more uniform. The structure has been used to build a Bipedal robot~\cite{gim2018biped},
achieving a 15\% decrease in the distance from origin to CoM (scaled to leg length), 
when compared to a reference robot with a serial kinematic chain~\cite{robotisop3}. Making assumptions such 
as walking only on flat ground has allowed for simplifying the control theory~\cite{kajita20013d}. Analogically, this can
be taken advantage of in the mechanical part. The NimbRo-OP2(X) robots~\cite{ficht2017nimbro}\cite{ficht2018nimbro} have adopted a five-DoF serial-parallel
leg structure with two 4-bar linkages in the sagittal plane and a serial chain for the lateral direction~(\figref{joints}.e.). The 4-bar pantographs 
constrain the leg orientation, keeping the foot parallel to the waist. When no tilting is present, the feet essentially stay
parallel to the ground. The actuation scheme is fundamentally different, where instead of the hip and knee joints, 
the angle is changed on the thigh and shank. Actuators can be placed at any pantograph axis and synchronised for increased performance.
A leg utilising this design can also be actuated completely at the hip, reducing the inertia~\cite{saab2017robotic}, 
as well as supplemented with a spring to reduce the torque requirements~\cite{han2011development}.

Biomimetics is a completely separate design concept, which inspires an increasing number of researchers. As humanoid robots
are supposed to mimic humans, it becomes natural to utilise not only the concept of a humanoid body plan, but the implementation 
of it as well. A musculoskeletal structure with biarticular actuation reduces the control bandwidth requirements and allows
for coordination and energy transfer between two joints to improve on efficiency~\cite{sharbafi2016new}\cite{roozing2019efficient}\cite{schumacher2020biarticular}. 
The usage of flexible elements and tendons improves on compliance and shock-loading of the limbs~\cite{schutz2017carl}.
As shown, the co-location of tendons with the skeleton---known as tensegrity---greatly reduces the stress on 
structural parts and allows for using lighter parts, resulting in improved dynamic performance~\cite{ananthanarayanan2012towards}. 
With multiple tendons going through several joints, minimising the friction or accounting for it becomes necessary~\cite{feldmann2014simulation}.
Completely biomimetic humanoids are still scarce. Kenshiro~\cite{nakanishi2012design}, and its successor Kengoro~\cite{asano2016human} are highly complex humanoids
with a human-like body structure including the skeletal structure and anatomically correct muscle arrangement. However, due to
the high complexity in the control and coordination of several 'muscles', the potential of such robots is yet to be explored.

\section{Actuation}

Designing a humanoid robot involves a holistic approach, where the structure of the robot is co-dependent on the chosen actuator and vice-versa.
From the mid 1980s, electrical actuators with high-ratio reducers have been the default choice, as they offer a good 
trade-off between torque, speed, and size. Apart from performance, using electrical actuators has benefits in simplifying the control, 
due to the nearly linear input-to-output relationship for control and convenient power storage and distribution systems.
The high transmission ratio required to provide the necessary torque has a drawback of a high-gain control regime.
As humanoid robots are required to interact more and more with their environment, force-feedback control becomes increasingly
relevant. This does not cope well with the rigid actuation of the high reduction setup, which interferes with 
the current/torque relationship and is not robust to shock-loading. This issue is adressed by introducing an elastic
element in the structure of the actuator, making it a \textit{Series Elastic Actuator}~(SEA), as used in NASA's Valkyrie~\cite{radford2015valkyrie}. The 
spring element serves several purposes: it absorbs impacts, stores energy, and controls the output torque through 
deflection using Hooke's law. WALK-MAN~\cite{tsagarakis2017walk} is equipped with the newest advances in SEA technology, utilising both active and 
passive adaptation~\cite{negrello2015modular}. Existing humanoids can be made compliant, by introducing
an elastic element on the output shaft of the actuator~\cite{martins2015polyurethane}.

The recent advances in Brushless DC~(BLDC) motor manufacturing and control allow for a new approach. Katz \cite{katz2018low},
cleverly combines the high torque density of a 'pancake' BLDC motor with a single-stage 1:6 planetary reduction embedded inside the stator bore.
As the commutation happens electronically using forward and inverse Park and Clarke transforms, it is possible to sense and control the output 
torque directly through the Q-axis current. In combination with an embedded impedance controller, these \textit{Quasi-Direct Drive}~(QDD) modules 
allow for robust, compliant and proprioceptive actuation with high peak power for dynamic locomotion. 
Although developed for the quadruped MIT Mini Cheetah~\cite{katz2019mini}, these actuators can be applied to bipeds 
as well~\cite{ramos2018facilitating}. Alterations of these actuators (e.g. longer stator, different reduction) with improved specifications 
are available on the market, making it a matter of time until compliant and dynamically capable humanoid robots become widespread.
Electrical motors are typically used for rotary movement; however, they can be used to produce linear motion through a 
translational (e.g. ball screw) mechanism~\cite{lohmeier2006leg} \cite{otani2013new}. They can also be produced as direct drive linear motors, which could be 
thought as simply 'unrolling' a rotary one on a guideway. 

Electrical motors in one form or another are limited in high power applications through magnetic saturation and thermal limits.
The higher the power requirements, the larger the motor. They are also quite fragile, as overloading them or operating
in sensitive environments (water, dirt) can quickly lead to them being destroyed. Hydraulics do not suffer from these limitations, 
as they offer high power density, scale well and can stop under a heavy load without damage even in harsh environments.
They are usually produced as pistons, where linear motion is performed by controlling the difference of incompressible liquid (usually oil) 
between two chambers separated by a movable cylinder. The achievable high-power specification of hydraulics is what attracted 
roboticists when building the first statically~\cite{kato1974information} and dynamically moving bipeds~\cite{playter1992control}. Currently, Boston Dynamics~\cite{bostondynamics}
is at the forefront of humanoid robotics using this technology. Their first humanoid PETMAN~\cite{nelson2012petman} achieved a human-like morphology,
range of motion, strength and walking. Its direct descendant ATLAS (along with ATLAS-DRC)~\cite{nelson2019petman} was built to investigate navigating 
in unstructured terrain. Highly impressive at the time, both robots showcased the limitation of the hydraulic system: the 
power storage and distribution were not as easy as with electrical systems, as the robots operated with a tether.
Also, every person in the vicinity of the robot was required to wear hearing-protection due to the generated noise.
This was resolved with ATLAS-Unplugged, where the hydraulic power unit was upgraded to produce variable pressure 
set-points, miniaturised and placed onboard of the robot. This redesign made ATLAS capable of full autonomy and much quieter operation, 
at the expense of a weight increase from \SI{152}{kg} to \SI{182}{kg}~\cite{nelson2019petman}. The new generation of ATLAS robots~\cite{atlasnew}
is the result of years of exploration and development of mobile hydraulic technology. By miniaturising the whole structure and 
components (especially the power unit), ATLAS became similar to a human in size~(\SI{1.5}{m}) and weight~(\SI{80}{kg}), 
also capable of elegant and powerful human-like movement including gymnastic routines~\cite{atlasgymnastics} and dancing~\cite{atlasdancing}.  

The path taken by Boston Dynamics shows that the entry point is steep and requires a synergy of interdisciplinary technical knowledge
to take full advantage of the benefits provided by hydraulic actuation. The technology is still unavailable commercially, meaning
that any entity interested in using a hydraulic system would face similar problems. During the DARPA Robotics Challenge (DRC) multiple 
competitors used the ATLAS platform, which was highly robust to not need maintenance for months, but repairs or upgrades had to be done 
by the manufacturer themselves---halting progress~\cite{johnson2017team}. Pairing the lack of access to technology with the 
unfamiliarity and inconvenience of working with it (noise, leakages) resulted in very few platforms built up to date. The most 
recent full-size humanoids include Hydra~\cite{kaminaga2016mechanism} and TaeMu~\cite{hyon2016design}, which differ in the actuation scheme. 
TaeMu employs the typical central pump scheme with servo valves to regulate the pressure across the robot, where
Hydra uses \textit{Electro-Hydrostatic Actuators}~(EHAs)~\cite{kaminaga2014development}, each with its own pump. The solution of Hydra
combines benefits from both electric and hydraulic actuation: backdrivability, impact resistance, controllability and power density, 
but is yet to be optimised for size and weight. In hydraulic systems, oil is still routed through flexible hoses, which reduces joint 
mobility and leads to leakages over time. Including oil pathways within the structural components of the robots is 
possible~\cite{li2019wlr}, but has to be done at the design level, adding more constraints to the process. As shown,
the process of designing and building even the few mentioned humanoids required custom solutions. This might reflect
the state of the options available on the market, more precisely the lack of standardised and modular solutions for hydraulically actuated robots.

\section{Sensing}

Similarly to humans, humanoids require the sense of self-movement and environment awareness. Over the years, multiple
sensing technologies have been developed and perfected. For body orientation, an Inertial Measurement Unit (IMU) composed
of an accelerometer, gyroscope and magnetometer is typically used. The readings are then fused to provide a full 3D-orientation of the body.
In combination with a model of the body plan and joint sensors, a full 6D estimation of self-movement can be obtained.
Joint position sensing is now usually done with high-resolution, hall-effect-based, magnetic encoders. The lack of friction 
in the sensing element (as in e.g. potentiometers) is a tremendous advantage, which mitigates the need for maintenance. 
Similarly, torque or force can be estimated by measuring the applied current to the joint, accounting for the transmission losses.
In combination with a Jacobian, end-effector wrenches can be estimated. This is sufficient for rough operation, but fine and 
detailed manipulation relies on accurate force-torque (F/T) sensor technology. A feedback loop designed to maintain a safe force, 
constantly applied to the manipulated object is necessary for a vast majority of applications such as opening a bottle or pushing a cart
as shown by ASIMO~\cite{shigemi2018asimo}. In locomotion, the stability assessment is also done using an estimation of the Center of Pressure (CoP)
obtained by F/T sensors. This puts a requirement on them being robust enough to not only carry the weight of the robot, but endure 
ground impact forces during walking. The drawback of currently available F/T sensors is that the measurements are 
done using transducers (e.g. strain gauges), with a direct coupling of tension induced by forces 
going through the sensor. This is also the reason for problems such as long-term deformations (creep) and overloading. 
Recalibration can prolong the usability of a sensor, unlike overloading caused by unpredictable shock impacts that simply destroy it. 
It is essential for new generations of F/T sensors to decouple the force from tension, and instead rely on contactless
displacement~\cite{kim2016novel} \cite{noh2016multi}, allowing to make the sensors more robust.  

For more elaborate multi-contact interactions, end-effector wrench estimation through F/T sensors 
or Jacobian methods is insufficient. Humans interact with the environment through skin, which allows for sensing 
various physical quantities such as temperature, humidity and pressure. It allows for distinguishing different 
types of contact, its amplitude as well as anticipating it through proximity sensing with hairs. There is a clear demand
for full-body robot skin \cite{atkeson2016happened} and although skin modules for robotics exist \cite{zou2017novel}, the 
technology is not mature enough to allow seamless integration with existing robots. The most advanced work in this regard is
HEX-o-SKIN~\cite{mittendorfer2011humanoid} \cite{mittendorfer2012integrating} incorporating force, temperature, proximity 
and acceleration sensors within an inexpensive, scalable, and robust package. The wide variety of applications developed with
these modules~\cite{mittendorfer2015realizing} \cite{kaboli2018active} \cite{leboutet2019tactile} \cite{kobayashi2019multi} 
\cite{rogelio2020plantar} underlines the importance of full-body skin. Full-body skin will also be essential in water-proofing
future humanoids for outdoor applications. 

\section{Manufacturing}

Manufacturing materials and techniques for robots matter as much as the technology utilised in them. Choosing the proper
materials for the structure impacts performance in terms of weight, rigidity, and maintenance. As seen in \tabref{robots}
metal alloys are the default choice in most humanoids, which can be attributed to the high rigidity, machinability and 
heat dissipation capabilities. For this, either subtractive (CNC-milling) or additive manufacturing (casting) is used, 
each with its benefits and drawbacks. The iCub humanoid~\cite{parmiggiani2012design} is a prime example of a 
robot constructed with a subtractive process, where a great majority of the 5000 parts is milled~\cite{metta2010icub}. 
The possible achievable shape is generally limited with milling, as the cutting tool needs to be able to execute a programmed path.
5-axis machines offer the highest fidelity, however enclosed parts and those with inner structures still need to be made out
of several sub-components. Casting can partially alleviate this problem, depending on the part complexity on a case-to-case basis. 
As shown on LOLA ~\cite{lohmeier2009humanoid}, casting allows to achieve quite sophisticated metal parts otherwise unachievable through 
milling, but requires a mold to be prepared for every part. In research, the low quantity requirements result in low 
cost-effectiveness of this method. 

Another option enhancing the production capabilities was added just recently, when consumer 3D printers became affordable and 
widespread~\cite{jones2011reprap}. 3D printing allows for fast production times of highly complex parts and an easy transition from
prototyping to production. In short time, several 3D-printed humanoid robots of various sizes have been designed and built \cite{lapeyre2014poppy}\cite{allgeuer2015child}\cite{wu2016hbs}\cite{ficht2017nimbro}\cite{ficht2018nimbro}.
The freedom presented through 3D-printing can be exploited either by an expert fine-tuning design details or by algorithms. 
Topology optimization can produce efficient shapes depending on user criteria (e.g. weight, rigidity) in the presence 
of constraints such as shape and material properties~\cite{junk2019additive}\cite{lohmeier2009humanoid}\cite{klemm2019ascento}.
There are still some issues when preparing the design files for printing. Due to the additive manufacturing process, the inner 
structure of the part is not uniform, unlike in subtractive manufacturing. This makes using the currently available stress 
analysis tools of CAD software ineffective, especially for deposition-based printing. Depending on the type and quality of the printer, the accuracy of
the finished print may vary greatly. The number of variables in the process is simply larger, such as the material used, the 
temperature, fill pattern or even orientation in which the part is printed. Ideally, the designed components should be disconnected 
from the production process, in 3D-printing however, they are very much dependent on it. Having in mind these issues and the 
limitations of the used printer, prototyping of a 3D-printed part is mostly done through iteration. Although there is continuous 
improvement in software and hardware capabilities of the printers, the deposition process is still limited. 
For production purposes, using a higher quality process such as Selective Laser Sintering~(SLS) can further increase the quality 
of the parts, making them more rigid and uniform in structre. The process is quite cost-effective for small production 
batches, after functional prototypes have already been produced. Aside from these polymer-based methods, 3D metal printing 
is already in advanced development~\cite{buchanan2019metal}, and has the possibility to completely revolutionise production of more 
than just robots. 

In addition to streamlining the production, 3D printing also facilitates the exchange between research institutions.
Developing and producing humanoid platforms for research requires technical and financial resources. The costs are multiplied 
with every new robot built as most components utilise customised and proprietary technology. Using standard, off-the-shelf,
commercially available components with open-sourced designs alleviates these issues in both development and maintenance costs~\cite{allgeuer2016igus}.
Obtaining a HUBO~\cite{park2007mechanical} requires hundreds of thousands of dollars, while the bill of materials for a 
similarly-sized open-source NimbRo-OP2(X) \cite{ficht2017nimbro}\cite{ficht2018nimbro}\cite{ficht2020nimbro} sums up to less than \SI{10}{\%} of that price. 
The base capabilities of open platforms are naturally less than those of established products, but they offer a level of 
customisation that can be taken advantage of. As researchers can reuse already developed solutions, they are not faced 
with the problem of 'reinventing the wheel'. Resources can then be spent on perfecting certain parts of a proven design instead of 
focusing on overcoming the initial hurdle of building a complete robot. By involving 3D printing in the process, the production 
and maintenance of robots no longer requires a separate machine shop for part production. A single 3D printer can not only reduce 
the production time, but completely alleviate the dependence from a part manufacturer. Broken parts could be swiftly improved, tested
and put on common servers to share with other users, which would lead to an accelerated development of humanoid robotics as a whole.

\section{Conclusions}

Humanoid hardware design requires extensive knowledge and experience in mechatronics to produce a capable platform. 
We have presented a glimpse at state-of-the-art technologies that contribute to humanoid robotics today as well as novelties
that will shape its future. The classical approach of highly rigid robots is gradually shifting towards compliant and dynamic ones 
for safer interaction with humans and more efficient locomotion. The discontinuation of ASIMO and the soaring popularity of ATLAS can 
be seen as a manifestation of this. Boston Dynamics are setting new, higher-than-ever standards for humanoid robotics, 
showing what can be achieved by breaking the typical design pattern. With easier access to rapid prototyping technologies, 
more full-scale humanoid robots are being built now than ever before. It will be exciting to see how all the technological 
advancements are utilised in future humanoid robots, and to what real tasks they can be applied to.

\section*{Compliance with Ethics Guidelines}
\textbf{Conflict of Interest} The authors declare that they have no conflict of interest.\newline
\textbf{Human and Animal Rights and Informed Consent} This article does not contain any studies with human or animal subjects performed by any of the authors.

\bibliographystyle{vancouver}
\bibliography{humanoidreview}

\end{document}